\documentclass{article}

\usepackage{arxiv}
\usepackage{color}
\usepackage[utf8]{inputenc} 
\usepackage[T1]{fontenc}    
\usepackage[hidelinks]{hyperref}       
\usepackage{url}            
\usepackage{booktabs}       
\usepackage{amsfonts}       
\usepackage{nicefrac}       
\usepackage{microtype}      
\usepackage{lipsum}		
\usepackage{float}
\usepackage{adjustbox}
\usepackage{makecell}
\usepackage{graphicx}
\usepackage[ruled,vlined]{algorithm2e}
\usepackage{multirow}
\usepackage{amsmath}
\usepackage{amsfonts}

\title{FabricNet: A Fiber Recognition Architecture Using Ensemble ConvNets}

\author{
	\href{https://orcid.org/0000-0001-7375-9040}{\includegraphics[scale=0.06]{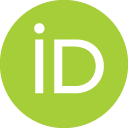}\hspace{1mm}Abu Quwsar Ohi} \\
	Department of Computer Science \& Engineering\\
	Bangladesh University of Business \& Technology\\ 
	Dhaka, Bangladesh \\
	\texttt{quwsarohi@bubt.edu.bd} \\
	\And
	\href{https://orcid.org/0000-0001-5738-1631}{\includegraphics[scale=0.06]{orcid.png}\hspace{1mm}M. F. Mridha} \\
	Department of Computer Science \& Engineering\\
	Bangladesh University of Business \& Technology\\ 
	Dhaka, Bangladesh \\
	\texttt{firoz@bubt.edu.bd} \\
	\And
	\And
	\href{https://orcid.org/0000-0001-9698-4726}{\includegraphics[scale=0.06]{orcid.png}\hspace{1mm}Md. Abdul Hamid} \\
	Department of Information Technology\\
	Faculty of Computing and Information Technology\\
	King AbdulAziz University\\ 
	Jeddah 21589, Kingdom of Saudi Arabia\\
	\And
	\href{https://orcid.org/0000-0003-2822-2572}{\includegraphics[scale=0.06]{orcid.png}\hspace{1mm}Muhammad Mostafa Monowar} \\
	Department of Information Technology\\
	Faculty of Computing and Information Technology\\
	King AbdulAziz University\\ 
	Jeddah 21589, Kingdom of Saudi Arabia\\
	\And
	Faris A Kateb \\
	Department of Information Technology\\
	Faculty of Computing and Information Technology\\
	King AbdulAziz University\\ 
	Jeddah 21589, Kingdom of Saudi Arabia\\
}

\date{}

\renewcommand{\headeright}{A.Q. Ohi et al.}
\renewcommand{\undertitle}{DOI: 10.1109/ACCESS.2021.3051980}

\begin{document}
\maketitle

\begin{abstract}
	Fabric is a planar material composed of textile fibers. Textile fibers are generated from many natural sources; including plants, animals, minerals, and even, it can be synthetic. A particular fabric may contain different types of fibers that pass through a complex production process. Fiber identification is usually carried out through chemical tests and microscopic tests. However, these testing processes are complicated as well as time-consuming. We propose FabricNet, a pioneering approach for the image-based textile fiber recognition system, which may have a revolutionary impact from individual to the industrial fiber recognition process. The FabricNet can recognize a large scale of fibers by only utilizing a surface image of fabric. The recognition system is constructed using a distinct category of class-based ensemble convolutional neural network (CNN) architecture. The experiment is conducted on recognizing 50 different types of textile fibers. This experiment includes a significantly large number of unique textile fibers than previous research endeavors to the best of our knowledge. We experiment with popular CNN architectures that include Inception, ResNet, VGG, MobileNet, DenseNet, and Xception. Finally, the experimental results demonstrate that FabricNet outperforms the state-of-the-art popular CNN architectures by reaching an accuracy of 84\% and F1-score of 90\%.
\end{abstract}

\keywords{
	Textile fiber recognition \and 
	Image processing \and 
	Convolutional neural network \and 
	Pattern recognition \and 
	Ensemble architecture
}

\section{Introduction}
\label{sec:introduction}
Textile fibers are the components that are used to construct fabrics. Commonly, the types of fibers are split into two categories: natural fibers and synthetic fibers. Natural fibers are extracted from environmental sources, whereas synthetic fibers are manufactured through machinery and chemical compounds. Such instances of natural fibers are silk, wool, cotton, etc. whereas, nylon, polyester, rayons, etc. are the example of synthetic fibers. Raw fibers are used to assemble yarns. A single yarn is assembled using one or more types of raw fibers. The yarns are further utilized to construct fabrics and particular garments.

Fiber recognition is the process of identifying raw fibers from fabrics, and it is widely used in different industrial applications. It is a well-applied method in fabric reverse engineering \cite{stylios1998engineering}.  Garment identification is also possible using fiber recognition systems since each garment type mostly requires fixed sets of fiber elements \cite{kampouris2016fine}. The fiber recognition system can also be implemented as fabric fault detection and garment inquiry systems \cite{ngan2011automated}.

\begin{figure}
	\center
	\includegraphics[width=0.5\linewidth]{{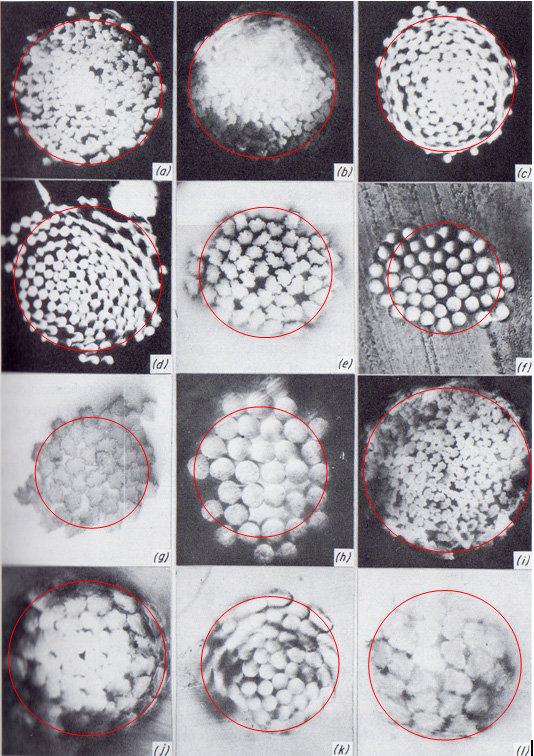}}
	\caption{Different yarn cross sectional shapes of particular fibers: Tenasco (a,b,i), Nylon (e,g,j,l), Viscose (f,k) and Terylene. The image is adopted from the work of Hearle et al. \cite{hearle1969structural}.}
	\label{fig:cross_section}
\end{figure}

Identifying raw textile fibers is considered difficult due to the complicated weave structure and aging of fibers \cite{goodway1987fiber}. Moreover, present fabrics pass through complex printing procedures that may alter the yarn structure. Classical biological methods, such as soaking, cleaning, heating, etc. are considered less effective in raw textile fiber identification. However, microscopic observations are proven to be more accurate in identifying raw textile fibers. The textile fiber recognition from manufactured fabrics is complicated since a single yarn (used to construct the fabric) can contain multiple textile fibers. In some cases, fabrics are mostly preprocessed. Moreover, microscopic observations may lead to false recognition, as numerous fibers can be used to generate a single textile yarn.

Generally, most systems identify textile fibers through microscopic cross-section images \cite{drobina2006application} and spectroscopic features \cite{sun2016classification,chen2019classification}. Fibers can be distinguished by microscopic cross-section images due to their unique geometrical properties \cite{ying2012single}. Figure \ref{fig:cross_section} illustrates different cross-section shapes of different types of yarns. Extricating cross-sectional shots is nearly impossible for industrial usage of textile fiber identification systems, as it requires a careful pre-processing and microscopes. Using cross-sectional images for recognizing textile fibers from fabrics is a critical approach in real-time industrial aspects. Hence, the cross-sectional investigation requires laboratories and is time-consuming. On the contrary, spectroscopy-based methods can be used for industrial purposes, but it is limited to recognizing only a single textile fiber from a fabric. Further, the method is not suitable for individual usage.

\begin{figure}
	\center
	\includegraphics[width=0.5\linewidth]{{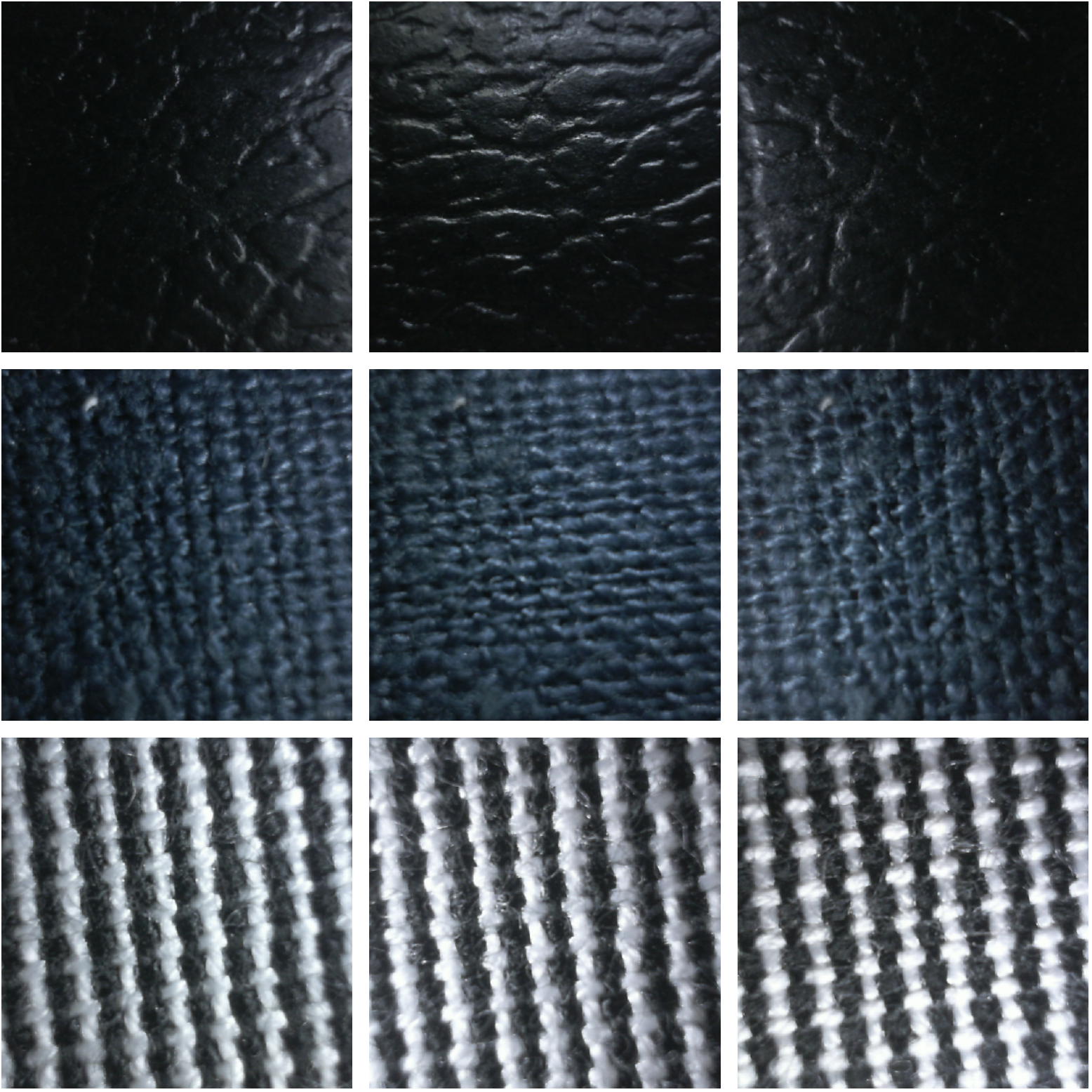}}
	\caption{The dataset contains fabric images in different light and orientations \cite{kampouris2016fine}. The first row illustrates fabric made of artificial leather. The second row illustrates the fabric made of silk. The third row contains images of fabric which is made of polyester and viscose (rayon).}
	\label{fig:fabric}
\end{figure}

The research endeavor's sole purpose is to overcome the information gathering complexities of the fiber recognition procedure. Smartphones and high definition cameras have made image capturing one of the most superficial attempts for information gathering procedures. Therefore, we introduce a novel architecture that can recognize the textile fibers by a fabric surface image. Because of the availability of cameras, our proposed textile fiber recognition process can be performed by individuals, and even by automated machines in a much more convenient and flexible way. Figure \ref{fig:fabric} represents some image samples that are used to conduct the training of our FabricNet architecture. The overall architecture performs CNN based image processing using an ensemble architecture. Since our proposed model recognizes textile fibers through fabric surface image, it can be widely applied for diverse industrial and individual applications for fault checking and authentication. Our process can further be used for textile fraud prevention and fault detection. The overall contribution of the research endeavor includes the following:

\begin{itemize}
	\item We exploit the surface image of fabric in recognizing textile fibers, as it is one of the easiest ways for image collection.
	
	\item We outline different categories of ensemble methods, and introduce a class-based ensemble architecture that receives downsampled image data through a head CNN architecture. In class-based ensemble architecture, every single ensemble memorizes only one class. Therefore, the accuracy of FabricNet architecture increases.
	
	\item We experiment with seventeen different implementations of famous image classification architectures, including Inception, ResNet, VGG, MobileNet, DenseNet, Xception, and CU-Net. Through the result analysis, we affirm that FabricNet architecture provides better accuracy.
\end{itemize}

The remainder of this paper is outlined as follows. Section \ref{sec:relatedworks} demonstrates the procedures that are modeled to identify textile fibers. Section \ref{sec:fabricnet} presents the motivation and the architectural fundamentals of the FabricNet. Section \ref{sec:experimentalresults} contains the experimental results that are performed to evaluate FabricNet architecture. Finally, Section \ref{sec:conclusion} concludes the paper.

\section{Related Works}
\label{sec:relatedworks}    
The textile fiber identification process can be divided into two types of tests: technical test and non-technical test \cite{khan2017review}. Before the advancement of microscopic tests, non-technical tests for textile fiber identification were mostly conducted. Buring test, soaking test, feeling, etc. are considered as non-technical tests. The drawback of the non-technical tests is its authenticity \cite{goodway1987fiber}.

Technical tests include the usage of microscopes and chemicals for the detection of textile fibers. Chemical tests include stain tests and solvent tests. However, the limitation of chemical tests is their inapplicability to separate multiple fibers. Therefore, chemical tests can not identify multiple fibers from fabrics \cite{corbman1983textiles}. Microscopic examinations are also considered as technical tests that involve identifying textile fabric using microscopes. Experts previously conducted microscope tests, and it proved to be more accurate. Nevertheless, synthetic fibers are mostly geometrically similar in microscopic view; therefore, specialists can sometimes find it challenging to distinguish fibers, even applying a microscope \cite{sawyer2008applications}. Figure \ref{fig:cross_section} illustrates an example of a microscopic test.

The usage of image recognition systems in fiber identification has been observed since late 1980 \cite{kinoshita1989determination}. The utilization of image processing for fiber recognition is observed since late 1990 that used Support Vector Machine (SVM) \cite{cortes1995support}. The SVM required a feature extraction method on which the effectiveness of the scheme was mostly dependent. The process also needed a microscopic cross-section view of fibers, which made the technique inefficient for real-time fiber recognition. The later research works conducted in the field of fiber recognition systems followed the same path. The purpose of the proposed architectures was to eliminate human interference from the technical microscopic fiber recognition procedure. Statistical analysis has also been introduced to identify fibers from cross-section images \cite{chiu1999fiber}. Nevertheless, the research work only classified two types of fibers, which is not suitable for real-world usage.

Neural networks became famous after the backpropagation learning method was introduced \cite{werbos1974beyond}. Neural network architecture is also introduced to identify fibers \cite{xu2007neural}. However, researches exploiting neural network architectures could only distinguish two types of fibers. Furthermore, neural networks do not fit in the image recognition process. Hence, neural network-based architectures are not suitable for recognizing a broad set of textile fibers.

CNN architectures started to rule in image identification problems through the successful breakthrough of AlexNet \cite{krizhevsky2012imagenet}. Till now, CNN architectures are considered the state-of-the-art image classification mechanism due to its robustness and ability to identify objects from a large set of targets. Also, CNN architectures perform auto feature extraction, eliminating the dependency from image feature extraction procedure \cite{cortes1995support}. CNN architecture has been introduced to identify fibers using cross-section microscopic images \cite{wang2017fiber}. The research was carried out by identifying seven types of fibers and achieved an acceptable accuracy level. Nevertheless, the method still depends on cross-section microscopic images, and the number of unique fibers is inadequate.

Apart from the same process of the automated identification of fibers from microscopic images, spectroscopy-based fiber identification methods have also been introduced \cite{sun2016classification,chen2019classification}. Nevertheless, the experiments of the research works were conducted with limited varieties of textile fibers, and the spectroscopy-based method can only identify a single fiber at once. The spectroscopy-based fiber identification process is also time-consuming and often requires an expert to position the spectroscope and calculate appropriate reading time correctly. 

The automated fiber identification has been less exploited, and most research endeavor is conducted to identify a single type of fiber at once. This type of classification is defined as multi-class classification. Furthermore, previously implemented architectures are time-consuming and require laboratory equipment to carry out the data collection (cross-section image or spectrograph extraction) and testing process. Therefore, mass testing and validating fibers of fabric often incur high cost. 

Feng et al. \cite{feng2019cu} have proposed a similar concept of converting the fiber recognition system into a multi-class classification problem. The authors presented a CNN-based ensemble architecture that contains an almost equivalent strategy of FabricNet architecture. The authors used a CNN architecture (defined as DFE module) to extract the fabric's necessary feature. Further, they transferred the feature vectors to a stack of ensemble network (referred to as CU module) containing three deep CNN architectures. Each of the models in the ensemble network also has inter-connectivity. Therefore, each of the models in the ensemble can be triggered by other models. Although the architecture is convincing, it suffers from overfitting issue due to enormous trainable parameters (approximately 82 million). Also, it requires extensive computations, which is about 3372 million floating point operations (FLOPs). The authors conducted evaluations in a closed source dataset and achieved 74\% accuracy. In comparison, our proposed architecture achieves superior accuracy with less trainable parameters and in fewer FLOPs.

\begin{table}
	\centering
	\caption{The table represents a detailed comparison of various domains of fiber recognition procedures. Among the different strategies, computer-vision-based method is fast, non-destructive, and requires no expertise. The table is a modified version, which was earlier presented by Z. Feng et al. \cite{feng2019cu}.}
	\label{tab:domain_survey}
	\begin{adjustbox}{width=\columnwidth,center}
		\begin{tabular}{|c|c|c|c|c|c|}
			\hline
			
			\textbf{Method} & \textbf{Accuracy} & \textbf{Requiring} & \textbf{Time} & \textbf{Destructive} & \makecell{\textbf{Identifiable fibre} \\ \textbf{number}} \\
			\hline \hline
			
			Physical & Expert's experience & Expert & > 5 min & Yes & Expert's experience \\ \hline
			
			Chemical & Expert's experience & Expert, reagent & > 10 min & Yes & Expert's experience \\ \hline
			
			Microscopic & Expert's experience & Expert, microscope & > 5 min & No & Expert's experience \\ \hline			
			
			Spectroscopy & > 0.8 & Infrared spectrometer & > 5 min & No & < 20 \\ \hline			
			
			Cut-section & > 0.7 & Optical aid & > 3 min & Yes & < 10 \\ \hline	
			
			\makecell{Computer Vision Systems\\ (CU-Net)} & 0.80 & \makecell{Camera, computation device} & \makecell{{< 1 s}\\{(3372 million FLOPs)}} & \textbf{No} & 50 \\ \hline
			
			\makecell{\textbf{Computer Vision Systems}\\ \textbf{(ours)}} & \textbf{0.84} & \makecell{\textbf{Camera,} \textbf{computation device}} & \makecell{\textbf{< 1 s}\\\textbf{(640 million FLOPs)}} & \textbf{No} & \textbf{50} \\ \hline
			
		\end{tabular}
	\end{adjustbox}
\end{table}

Table \ref{tab:domain_survey} represents a comparison of various types of fiber recognition architectures that also consider computer vision framework. The present vision-based architecture is faster, accurate, non-destructive, and do not require any expertise to identify textile fibers. By analyzing the table, it can be concluded that the computer vision based strategy is more prominent than all the previous domains introduced till now. 

We present an architecture that operates over a more user-friendly information extraction process. It requires a close-shot surface image of fabric to identify the raw textile fibers. We argue that close-shot surface images contain proper fabric surface properties that are enough to identify textile fibers. Furthermore, in the current new media age, image capturing is one of the simplest functions due to the well-spread of smartphones and high-definition cameras. Therefore, the architecture changes the process of textile fiber extraction from laboratories to individuals and industries. Moreover, we achieve a satisfactory recognition accuracy that is most suitable for industrial purposes.

The proposed FabricNet architecture can identify multiple fibers at once. Therefore, the classification process of the model is multi-label. The overall architecture uses CNN containing a class-based ensemble that can individually recognize a specific fiber. We investigate with various deep learning frameworks and determine that Xception \cite{chollet2017xception} performs optimally amongst the existing CNN architectures. We further adapt the structure of the Xception architecture and assign our ensemble strategy. We name the revised version of the Xception architecture as FabricNet.

\section{FabricNet}
\label{sec:fabricnet}

\subsection{Motivation}
\label{sec:motivation}

In neural network architecture, ensemble methods \cite{hansen1990neural} combine multiple sub-models to obtain better accuracy. Ensemble architectures have been proven to be less prone to overfitting and generate more accurate results than basic models \cite{tao2019deep}. Furthermore, present ensemble architectures have been able to identify complex spatial information from image patterns \cite{xu2020multi}. Therefore the ensemble methods are being implemented in different scopes, including geospatial land classification \cite{minetto2019hydra}, face recognition \cite{ding2017trunk}, image segmentation \cite{tan2019evolving}, and so on. The process of ensemble methods can be thought of as a particular situation where a group of people will always make better decisions than a single person \cite{bishop1995neural}. Dietterich \cite{dietterich2000ensemble} pointed out three reasons for which the ensemble architecture may work better than traditional architectures: 1) training phase may not contain sufficient data to train the single best classifier; 2) a single algorithm may fail to converge to the global optimum, but an ensemble starting from distinct points could lead to a better approximate global optimum; 3) the space being searched may not contain any optimum position, but an ensemble may lead this space for a better optimum position.

In deep learning architecture, three types of ensemble architectures are mostly encountered, a) stacked ensembles, b) weight average ensembles, and c) class-based ensembles. In stacked ensemble architecture, multiple sub-models receive input data and flows the data stream to a final learning model that generates the results. Mathematically, the stacked ensemble can be presented as,
\begin{equation}
	\begin{split}
		&E(x) = f(e_1(x), e_2(x), ..., e_n(x))
	\end{split}
\end{equation}
\begin{equation}
	\begin{split}
		Where,& \nonumber\\
		& x = \text{Input data}\\
		& E(x) = \text{Stacked ensemble model}\\
		& e_i(x) = \text{Ensemble sub-models}\\
		&n = \text{Number of ensemble sub-models}\\
		&f(x) = \text{The final learning model}
	\end{split}
\end{equation}

In weight average ensemble architecture, the results of multiple models are calculated separately, further combined through weight multiplication calculations to perform final prediction \cite{acar2015effect}. Mathematically, the weight average ensemble can be presented as,
\begin{equation}
	\begin{split}
		&E(x) = argmax(w_1 \times e_1(x), w_2 \times e_2(x), ..., w_n \times e_n(x))
	\end{split}
\end{equation}
\begin{equation}
	\begin{split}
		Where,& \nonumber\\
		& x = \text{Input data}\\
		& E(x) = \text{Weight average ensemble model}\\
		& e_i(x) = \text{Ensemble sub-models}\\
		& w_i = \text{Weights for each ensemble models}\\
		&n = \text{Number of ensemble sub-models}\\
	\end{split}
\end{equation}

In class-based ensemble architecture, the number of ensemble models is equal to the number of classes \cite{al2015recurring}. Each of the ensemble models only learns to identify a specific category. The class-based ensemble architecture can be derived as,
\begin{equation}
	\begin{split}
		&E(x) = argmax(e_1(x),e_2(x), ..., e_n(x))
	\end{split}
\end{equation}
\begin{equation}
	\begin{split}
		Where,& \nonumber\\
		& x = \text{Input data}\\
		& E(x) = \text{Class-based ensemble model}\\
		& e_i(x) = \text{Ensemble sub-model for $i^{th}$ class}\\
		&n = \text{Number of output classes}\\
	\end{split}
\end{equation}

\begin{figure}[!h]
	\center
	\includegraphics[width=\linewidth]{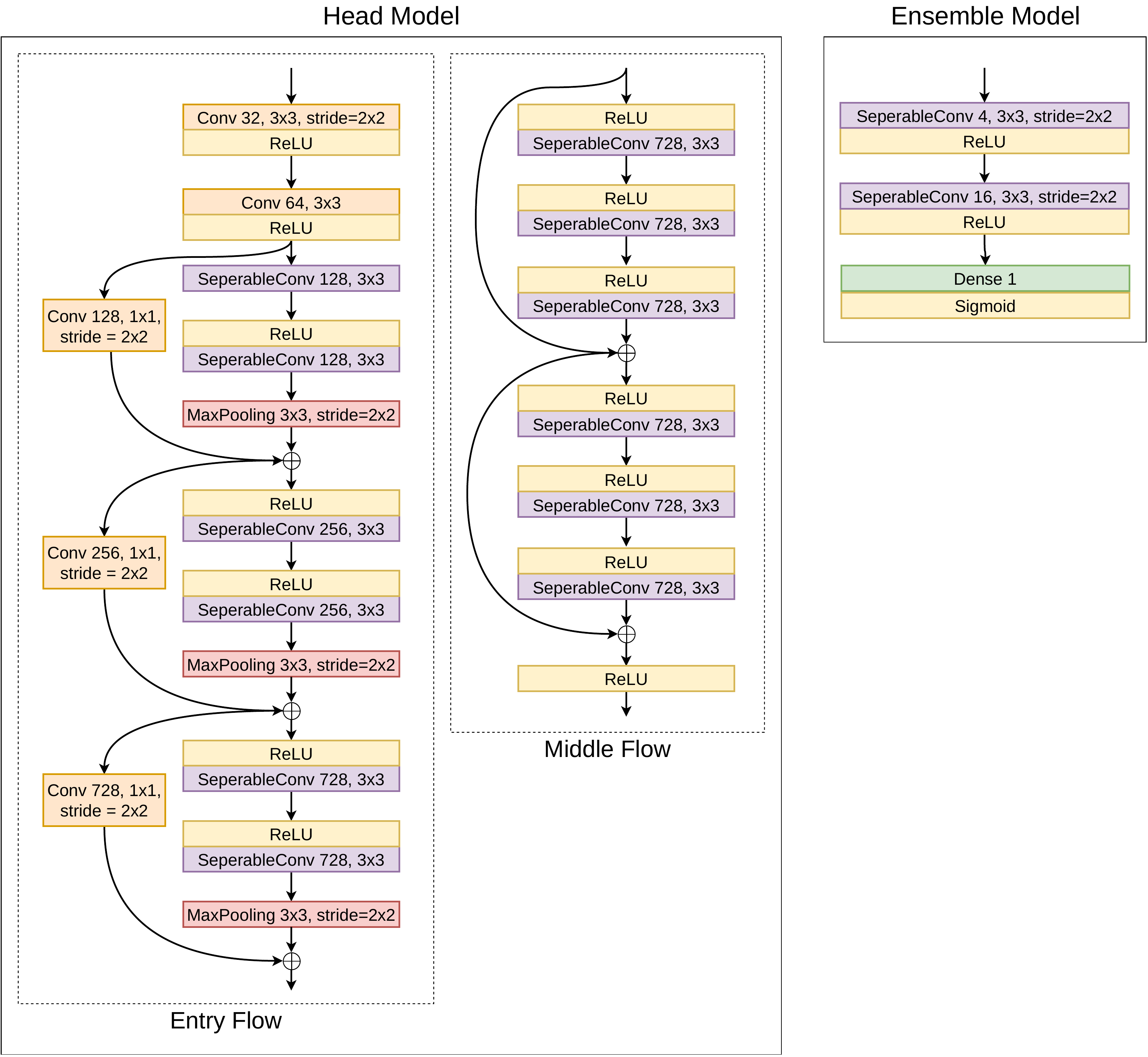}
	\caption{The figure illustrates the head and ensemble model of the FabricNet. The entry flow network recieves input image, and the processed data is passed to the middle flow network. It further forwards the processed data to the multiple ensemble models. Each of the convolutions is followed by a batch normalization \cite{ioffe2015batch} layer, not illustrated in the image.}
	\label{fig:model}
\end{figure}

\begin{figure}[!h]
	\center
	\includegraphics[width=0.5\linewidth]{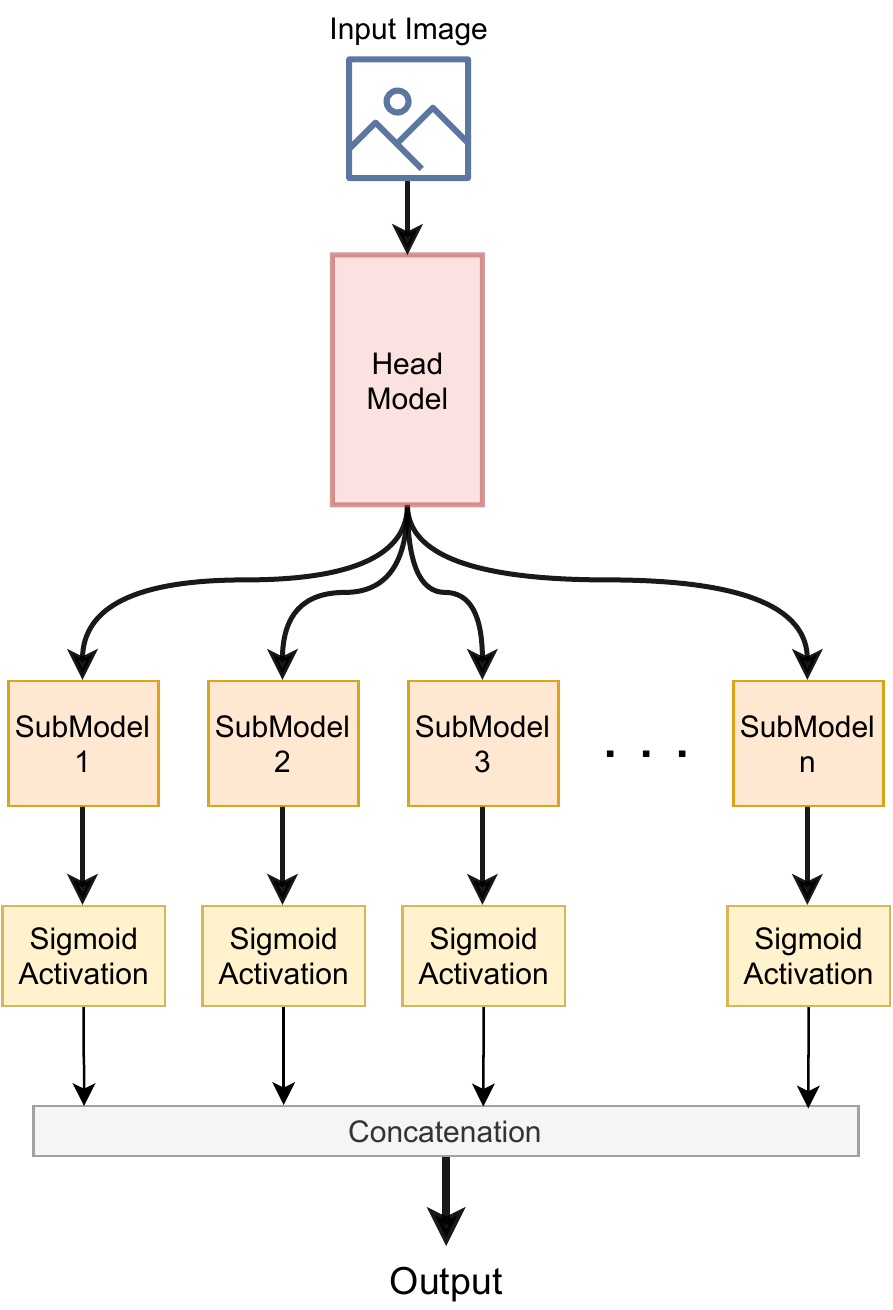}
	\caption{The figure depicts the architectural strategy of the FabricNet model. Inputs flow through the head model, which is further passed through the class-specific ensemble submodels. Submodels contain minimal trainable parameters to avoid overfitting and reduce computational complexity.}
	\label{fig:ensemble}
\end{figure}

Inspired by the performance boost of the ensemble architectures, we develop an ensemble architecture that performs mostly similar to class-based ensemble architecture. Nevertheless, our proposed architecture is slightly different than the class-based ensemble architecture. Mathematically, the proposed architecture can be derived as follows,
\begin{equation}
	\begin{split}
		&E(x) = argmax(e_1(f(x)),e_2(f(x)), ..., e_n(f(x)))
	\end{split}
	\label{eq:fabricnet}
\end{equation}

\begin{equation}
	\begin{split}
		Where,& \nonumber\\
		& x = \text{Input data}\\
		& E(x) = \text{FabricNet model}\\
		& e_i(x) = \text{Ensemble sub-model for $i^{th}$ class}\\
		&n = \text{Number of output classes}\\
		&f(x) = \text{Head model}
	\end{split}
\end{equation}

Instead of directly passing the ensembles' inputs, the proposed architecture uses an auxiliary feature extractor function defined as the head model. The head model only passes the relevant feature embeddings to the ensemble models and reduces the full dependency over the ensemble models. As a fabric can be constructed using multiple fibers, the FabricNet must output multiple classes at once, often acknowledged as multi-label classification. Therefore, the $E(x)$ function of equation \ref{eq:fabricnet} may return multiple outputs at once. Also, the number of ensembles $n$ is kept equal to the number of target classes. Furthermore, each ensemble model $e_i(x)$ specifically learns to identify a particular class. Therefore, individual ensemble models can approach an optimal state to recognize a specific class, ignoring other classes. This individuality may cause improving the accuracy of the FabricNet architecture.

\subsection{Architecture}
\label{sec:architecture}

The overall architecture of FabricNet is segmented into two parts: a head model and ensemble models. The head model directly fetches the input images and generates lower dimension embeddings. The embeddings derived by the head model are passed to the ensemble models. Each of the ensemble models is assigned to identify only a single type of fabric fibers or class, and each class's prediction is independent of the other class-based ensemble. Therefore, the number of ensemble models must be similar to the number of possible categories. Figure \ref{fig:ensemble} illustrates a block diagram of the FabricNet architecture. Using the head model poses some advantages, considering the usual class-based model. Generally, higher parameters in a CNN architecture may often cause overfitting. Passing the input through a head model causes a reduction of irrelevant features. Furthermore, it also reduces the number of trainable parameters significantly. On the whole, implementing our suggested ensemble architecture leads to the following benefits: 
\begin{itemize}
	\item Each ensemble only extracts the knowledge required to recognize a specific class. This results in acquiring an approximate optimal position for each particular class, which causes superior accuracy.
	\item Using the head model significantly reduces the number of parameters required for each ensemble model. This substantially reduces the FLOPs and overfitting \cite{maida2016cognitive}.
	\item Apart from the general class-based ensemble, our proposed ensemble structure can be deeper with reduced computational complexity (due to head convolution non-parallelism). 
\end{itemize}

Instead of implementing a new baseline architecture for the head model, we modify the existing CNN architectures. Through our investigation (illustrated in Figure \ref{fig:f1_baselines}), we affirm that Xception architecture performs superior to the currently existing popular architectures. Therefore, we fuse the ensemble methodology in Xception architecture. The Xception model counterfeits the basic properties of VGG \cite{simonyan2014very} and Inception \cite{szegedy2015going} network. The model utilizes a shorter kernel of the size of 3 and performs depthwise separable convolutions. The depthwise separable convolutions of the Xception architecture are performed by a depth-wise convolution followed by a pointwise convolution (in Inception architecture, the order is reversed). This strategy is based on a hypothesis that spatial feature extraction and channel-wise feature extraction procedure can be decoupled. Meanwhile, this decoupling has a huge advantage in reducing the required parameters of a convolutional layer. Xception further implements residual identity maps \cite{he2016deep} in different layers to resolve the issue of vanishing gradient. 

Xception architecture contains three types of data-flow networks. The entry flow performs depthwise separable convolutions, followed by a maxpool layer. The entry flow is followed by nine similar middle flow networks that perform depthwise separable convolution. Finally, the exit flow network performs a similar computation sequence to the entry flow, followed by a global average pooling. The model is substantially deep (126 layers), and it is often validated that deeper networks are better \cite{simonyan2014very}. However, as we search for an optimal head model that will also contain ensemble models, we have to avoid over-parameterizing the head model. Over-parameterization may cause overfitting and also increase the FLOPs of the architecture.

Hence, we expel unnecessary blocks from the Xception architecture, which does not boost the prediction accuracy. While expelling the exit flow and some middle flow networks of the Xception model, we found a minimal fluctuation of F1-score and AUC score (reported in Figure \ref{fig:xception_flows}). The metrics' steadiness indicates that most of the lower layers are not necessary (for the experimental dataset), and they can be removed. Therefore, we only adopt the entry flow and two middle flow stacks of the Xception architecture as our head model. Also, removing the lower portion of the Xception architecture reduces trainable parameters by more than 80\%. A full investigation is reported in the Result Analysis section (Section \ref{sec:resultcomparison}).

The architectural specifications of the head and ensemble models of the FabricNet is reported in Figure \ref{fig:model}. Moreover, Figure \ref{fig:ensemble} illustrates the overall flow of the model. We implemented separable convolutions in the ensemble model to keep the learnable parameters limited. Each ensemble model contains a single dense node with a sigmoid activation function that works as the final activation for each category. The available outputs of each ensemble are further concatenated to produce the final output of the FabricNet model. We broadly analyze and discuss our FabricNet architecture findings in the Result Analysis (Section \ref{sec:resultcomparison}), particularly the ensemble architecture.

As the FabricNet architecture performs a multi-label classification task, the final output contains a sigmoid activation function. Nam et al. has suggested that cross-entropy loss is the best choice for multi-label classification tasks \cite{nam2014large}, that is defined as follows,
\begin{equation}
	\begin{split}
		&L_{CE}(y, o) = - \left( \sum\limits_{l} \; (y_l \; log \; o_l) \; - \; (1 \; - \; y_l) \; log(1\; -\; o_l) \right)
	\end{split}
	\label{eq:loss}
\end{equation}
\begin{equation}
	\begin{split}
		Where,& \nonumber\\
		&L_{CE} = \text{Loss function}\\
		&o_l = \text{Prediction for label $l$}\\
		&y_l = \text{Target for label $l$}\\
	\end{split}
\end{equation}

Therefore, the model is trained using the aforementioned cross-entropy loss function.

\section{Experimental Results}
\label{sec:experimentalresults}

\subsection{The Fabrics Dataset}
\label{sec:thefabricsdataset}

\begin{figure}[!h]
	\center
	\includegraphics[width=0.5\linewidth]{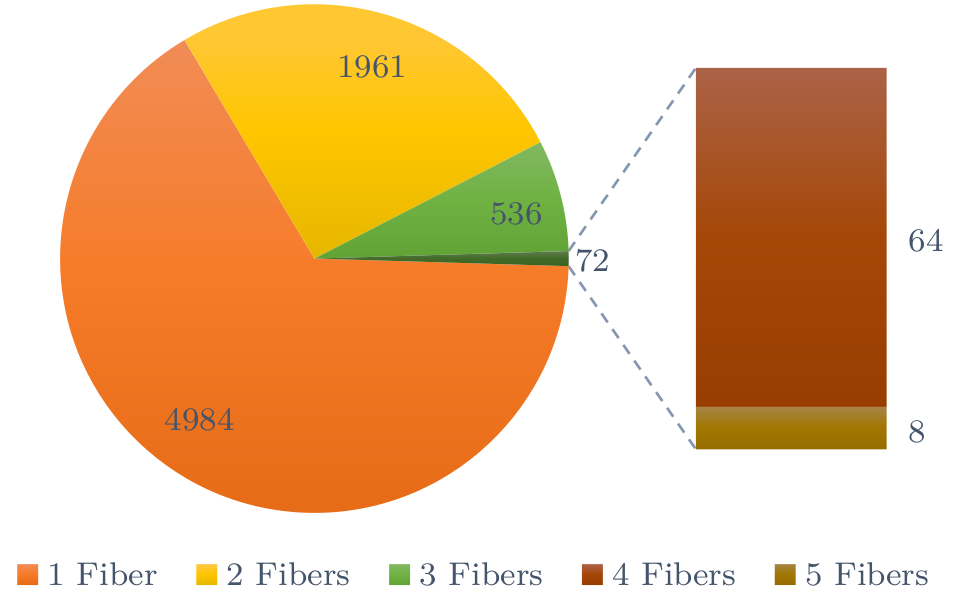}
	\caption{A pie chart that represents the number of images belonging to the number of fiber classes.}
	\label{fig:datasetcount}
\end{figure}

As this is a prior work investigating multiple fibers, we currently found only one dataset suitable for the experiment. The fabric dataset contains around 8000 images of different fabrics, and garments \cite{kampouris2016fine}. However, we found a total of 7553 images suitable for the experiment. Although the original work conducted using the dataset contained only 2000 images of fabric surfaces, the current repository contains an increased number of surface images. The dataset contains images with various lightning and orientation to make the recognition process more challenging. Figure \ref{fig:fabric} contains an example of the images. Further, Each of the fabric images contains one or more classes. The classes are the fibers that are used to construct the fabric. These fabrics contain a total of fifty types of fibers that constituted the fabrics. Figure \ref{fig:datasetcount} illustrates the class distribution for the images of the dataset. The dataset collectors only attempted to identify fabrics that were not blended (contained a single fiber) \cite{kampouris2016fine}. Therefore, the collectors were able to identify nine types of non-blended fabrics from the dataset. 

\subsection{Evaluation Metrics}
\label{evaluationmetrics}
To evaluate and compare the results, three evaluation metrics have been used, which are presented as follows:\\

\textbf{Accuracy:} Accuracy is the simplest form of evaluation. It formally defines the ratio of correct predictions over total experiments. In multi-label classification, we consider a single prediction is accurate if all of the classes are correctly guessed. Accuracy can be defined as follows,
\begin{equation}
	Accuracy = \frac{\text{Number of correct predictions}}{\text{Total number of predictions}}
	\label{eq:acc}
\end{equation}

\textbf{Precision:} Precision is also named as the positive predictive value (or true positive rate) of a system that reports the ratio of correctly predicted positive cases over total predicted positive cases. It can be represented as,
\begin{equation}
	Precision = \frac{TP}{TP+FP} 
	\label{eq:pre}
\end{equation}

$\mathbf{F_1}$\textbf{-Score:} $F_1$ score represents the weighted average of precision and recall. By choosing the weight value as 2, the $F_1$ score can be presented as,
\begin{equation}
	F_{1}Score = \frac{2 \times Precision \times Recall}{Precision+Recall}
	\label{eq:f1}
\end{equation}

\textbf{AUC Score}: Area under the curve (AUC) and receiver operating characteristics (ROC) curve defines how well a model converges towards distinguishing classes accurately. In general, the metric generates a curve, where AUC represents the area under the ROC curve. In overall experiments, we use 200 thresholds to discretize the ROC curve.

\textbf{FLOPs}: Floating point operations (FLOPs) measures the number of arithmetic operations required to execute a single instance of a deep learning model. Models requiring higher FLOPs have higher time complexity. 

Accuracy, Precision, F1-score, and AUC score generate results in the range [0, 1], whereas higher score points better performance of a system. Hence, we use the metrics mentioned above to prove the effectiveness of our model. Moreover, we use FLOPs to measure the time complexity of each model.

\subsection{Experimental Setup}
\label{sec:experimentalsetup}

The evaluation architectures were implemented using Tensorflow \cite{abadi2016tensorflow}, Keras \cite{chollet2015keras}, scikit-learn \cite{pedregosa2011scikit}, and NumPy \cite{walt2011numpy}. To lessen the architecture's bias and correctly measure each architecture's accuracy, k-fold cross-validation is performed \cite{mosteller1968data}. All of the evaluations are conducted by selecting the value of $k = 4$. Therefore, the dataset is split into 50\%-25\%-25\% train, validation, and test subsets. The reported measurements are the best performance on the validation set for a particular fold, further evaluated on the unseen test set. Each architecture is trained using $batchSize=128$ with a maximum epoch limit of $100$. With a learning rate of $0.001$, Adam optimizer is used to train each of the architectures. All of the tested architectures are initialized with ImageNet trained weights, and they are further trained on the Fabrics dataset. The input image is in the shape of $120\times120\times3$. As the training dataset is small, we implemented image augmentation considering some common augmentation process that includes brightness change, contrast change, zooming, cropping, and channel shifts. We did not perform any geometrical distortions as it may change the texture pattern of the fabric weave. Every exhibited result is calculated as the mean and standard deviation for three runs for each fold (total $3\times4$ runs). 

\begin{figure}[H]
	\center
	\includegraphics[width=\linewidth]{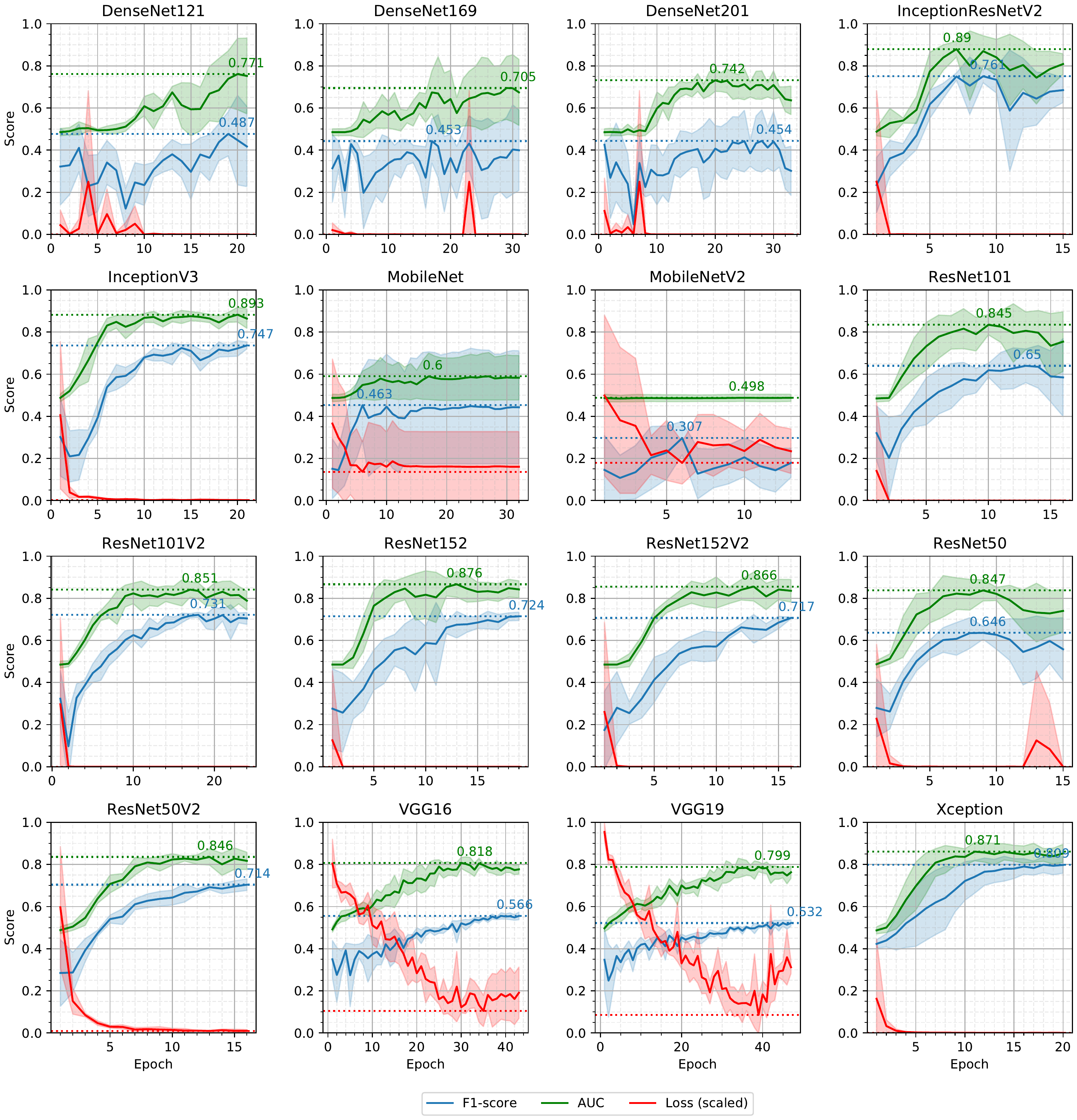}
	\caption{Each graph represents the loss (min-max normalized), AUC, and F1-score on the validation dataset calculated on the existing DCNN architectures' training procedure. The horizontal axis represents the training epochs, while the vertical axis represents the metric score. The train and test scores of the corresponding architectures are reported in Table \ref{tab:acc}. Zoom in for a better view.}
	\label{fig:f1_baselines}
\end{figure}

\begin{figure}[H]
	\center
	\includegraphics[width=\linewidth]{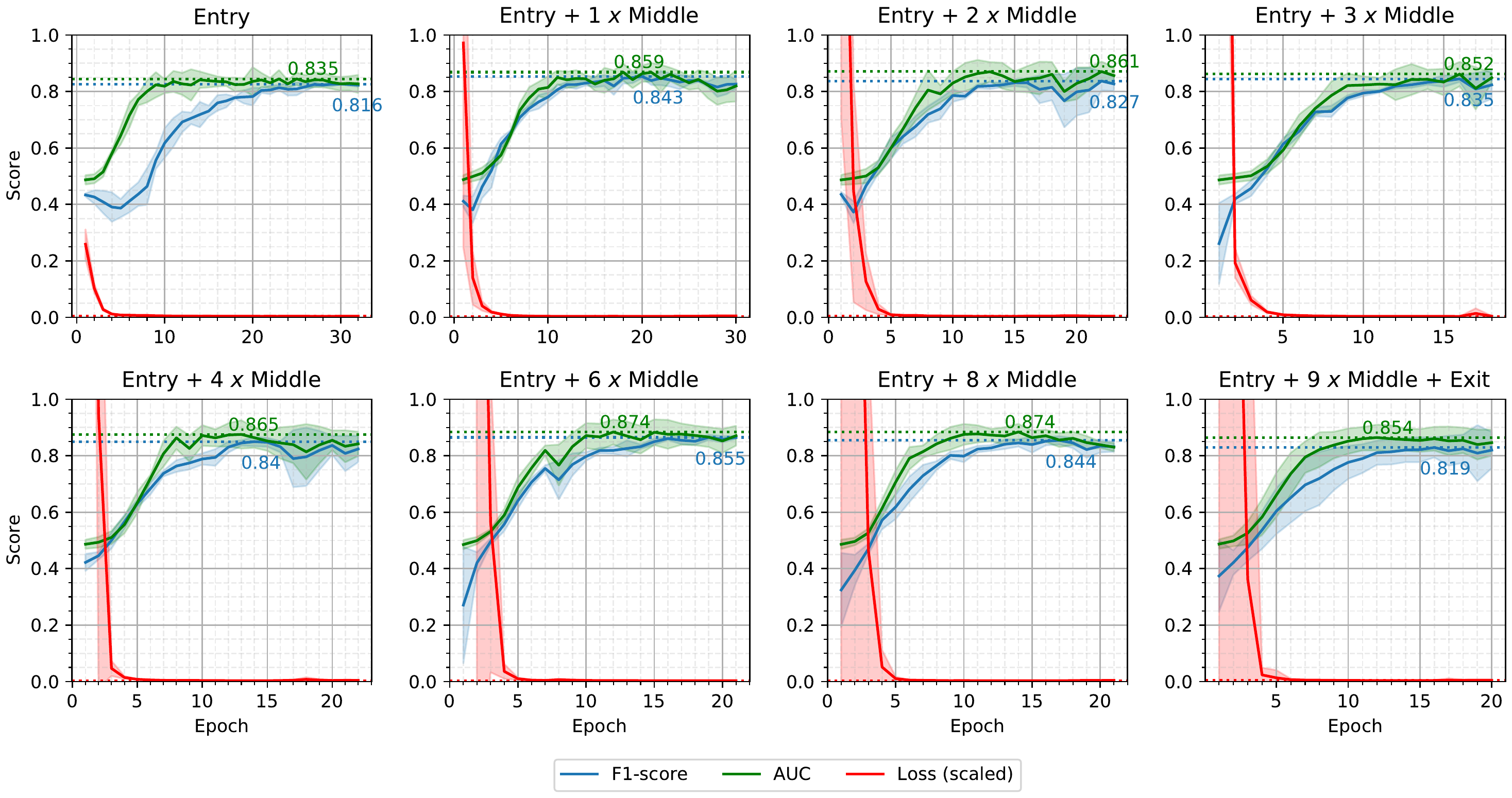}
	\caption{Each graph represents the loss, AUC, and F1-score on the validation dataset calculated using different Xception architecture setups. Entry, middle, and end define the three types of network flows of Xception architecture. The number of middle flows is indicated by multiplication. The `entry$+2\times$middle' design is used as the head model.}
	\label{fig:xception_flows}
\end{figure} 

\begin{figure}[H]
	\center
	\includegraphics[width=\linewidth]{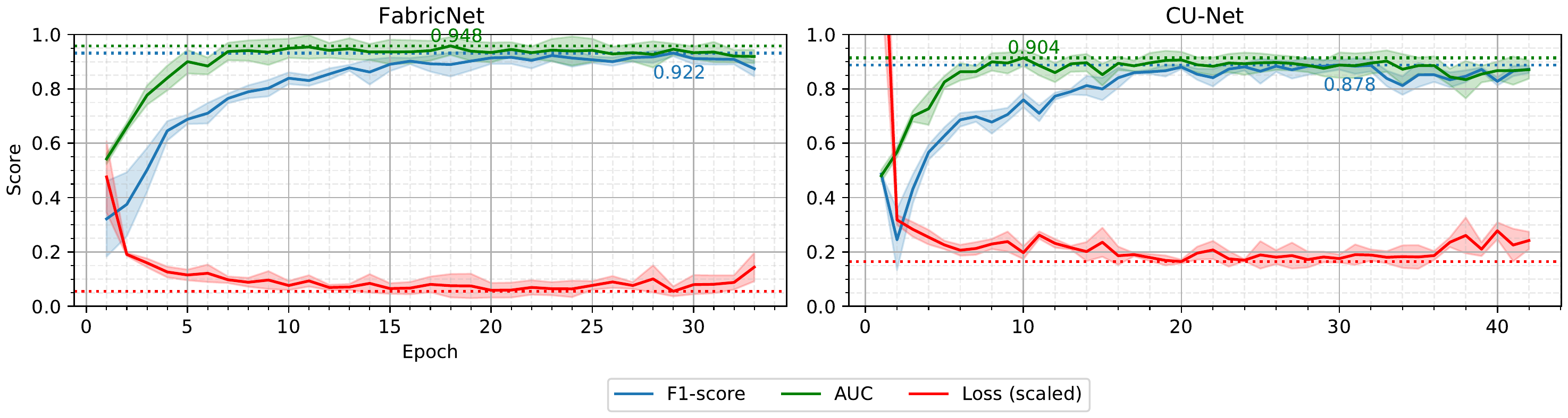}
	\caption{The figure illustrates a comparison of FabricNet architecture with CU-Net architecture. Both architectures are based on ensemble strategy. However, FabricNet architecture acquires higher performance as it contains class-specific models in the ensemble.}
	\label{fig:fabricnet_flows}
\end{figure}

\subsection{Result Analysis}
\label{sec:resultcomparison}

As fabric surface images are less exploited for fabric fiber recognition, we compare the FabricNet architecture with various computer vision based architecture. Also, we include the existing architecture CU-Net \cite{feng2019cu} that also operates over fabric surface image. We have implemented the DenseNet baseline for the CU-Net architecture.

Figure \ref{fig:f1_baselines} illustrates the F1-scores obtained on the validation set while training the existing DCNN frameworks in the fabric dataset. The graphs report that Xception architecture achieves a better result on the validation set. VGG architectures do not perform adequately in the framework, mostly due to the vanishing gradient problem. On the contrary, ResNets solve the vanishing gradient problem by implementing residual identity maps. Yet they don't acquire satisfactory results mostly due to overfitting. InceptionResNet architecture achieves better results due to the proper integration of residuals and inception blocks. DenseNets require fewer parameters than Xception architecture; still, the idea of the shorter connection from input and output does not help to achieve better performance. MobileNet architectures require the least number of parameters than the other implemented models. Nevertheless, they fail to perform fiber recognition at an acceptable rate. This low F1-score of MobileNets may indicate that the issue is not with training parameters, rather than a necessity of optimal network architecture.

\begin{table}[!h]
	\caption{\label{tab:ensemble_archi} The table illustrates the F1-score of the validation and test dataset corresponding to the different ensemble architectures. Each layer is represented as `$\{Sx,y,z\}$', where `$x$' is the number of filters, `$y$' is the kernel size, and `$z$' is the stride. `$S$' defines a depthwise separable convolution layer. Each convolution is followed by batch normalization and ReLU activation function. Trainable parameters (for each ensemble) are presented in thousand.}
	
	\centering
		\begin{tabular}{|c|c|c|c|}
			\hline
			\textbf{Architecture} & \makecell{\textbf{Parameters}\\\textbf{(thousand)}} & \textbf{F1-score (val)} & \textbf{F1-score (test)} \\
			\hline \hline
			
			\{S64,3,2\},\{S64,3,2\} & 58 & 0.883±0.01 & 0.80±0.01 \\ \hline
			
			\{S32,3,2\},\{S64,3,1\} & 33 & 0.884±0.01 & 0.828±0.01 \\ \hline
			
			\{S16,3,2\},\{S32,3,1\} & 19 & 0.884± 0.0 & 0.874±0.01 \\ \hline
			
			\{S8,3,2\},\{S8,3,1\} & 12 & 0.893±0.02 & 0.887±0.02 \\ \hline
			
			\textbf{\{S4,3,2\},\{S16,3,2\}} & \textbf{9} & \textbf{0.922±0.002} & \textbf{0.902±0.01} \\ \hline
			
			\{S2,3,2\},\{S16,3,2\} & 8 & 0.891±0.01 & 0.889±0.02 \\ \hline
			
			\{S4,3,2\},\{S16,3,2\}\{S16,3,2\} & 9 & 0.90±0.01 & 0.898±0.02 \\ \hline
			
		\end{tabular}
\end{table}

\begin{table}[H]
	\centering
	\caption{The table reports a comparison between the FabricNet and other well-performing architectures on train validation and test dataset. The Parameters column represents the total number of trainable parameters required for each model. \\}
	\label{tab:acc}
	\begin{adjustbox}{width=\linewidth,center}
		\begin{tabular}{|c|c|c|c|c|c|c|c|c|c|c|c|c|c|c|c|c|c|c|}
			\hline
			\multirow{2}{*}{\textbf{Architecture}} &
			\multirow{2}{*}{\makecell{\textbf{Parameters}\\\textbf{(million)}}} & 
			\multirow{2}{*}{\makecell{\textbf{FLOPs}\\\textbf{(million)}}} & 
			\multicolumn{4}{c|}{\textbf{Train}} &
			\multicolumn{4}{c|}{\textbf{Validation}} &
			\multicolumn{4}{c|}{\textbf{Test}} \\ \cline{4-15}
			
			& & & \textbf{Precision} & \textbf{Accuracy} & $\mathbf{F_1}$ \textbf{-Score} & \textbf{AUC} & \textbf{Precision} & \textbf{Accuracy} & $\mathbf{F_1}$ \textbf{-Score} & \textbf{AUC} & \textbf{Precision} & \textbf{Accuracy} & $\mathbf{F_1}$ \textbf{-Score} & \textbf{AUC} \\ \hline \hline
			
			MobileNetV2 \cite{sandler2018mobilenetv2} & 3 & 96 & 
			0.964±0.0 & 0.851±0.01 & 0.957±0.003 & 0.998±0.0 &
			0.718±0.235 & 0.394±0.077 & 0.307±0.069 & 0.498±0.007 &
			0.553±0.01 & 0.494±0.01 & 0.418±0.01 & 0.503±0.01  \\ \hline
			
			VGG19 \cite{simonyan2014very} & 143 & 5533 & 
			0.762±0.015 & 0.658±0.01 & 0.641±0.029 & 0.962±0.012 &
			0.669±0.005 & 0.574±0.006 & 0.532±0.005 & 0.799±0.014 &
			0.696±0.01 & 0.585±0.01 & 0.528±0.01 & 0.809±0.01  \\ \hline
			
			VGG16 \cite{simonyan2014very} & 138 & 4356 & 
			0.79±0.001 & 0.746±0.01 & 0.689±0.014 & 0.973±0.006 &
			0.686±0.017 & 0.585±0.0 & 0.566±0.005 & 0.818±0.016 &
			0.687±0.01 & 0.583±0.01 & 0.561±0.01 & 0.844±0.01 \\ \hline
			
			DenseNet201 \cite{huang2017densely} & 20 & 1134 & 
			0.724±0.086 & 0.765±0.01 & 0.594±0.136 & 0.839±0.085 &
			0.603±0.084 & 0.528±0.007 & 0.454±0.138 & 0.742±0.043 &
			0.809±0.01 & 0.662±0.01 & 0.676±0.01 & 0.85±0.01  \\ \hline
			
			DenseNet121 \cite{huang2017densely} & 8 & 783 & 
			0.787±0.029 & 0.79±0.01 & 0.697±0.049 & 0.955±0.009 &
			0.644±0.092 & 0.506±0.077 & 0.487±0.095 & 0.771±0.159 &
			0.778±0.01 & 0.674±0.01 & 0.677±0.01 & 0.918±0.01 \\ \hline
			
			DenseNet169 \cite{huang2017densely} & 14 & 903 & 
			0.72±0.055 & 0.782±0.01 & 0.583±0.103 & 0.834±0.143 &
			0.604±0.065 & 0.532±0.059 & 0.453±0.094 & 0.705±0.139 &
			0.85±0.01 & 0.71±0.01 & 0.753±0.01 & 0.918±0.01  \\ \hline
			
			ResNet152 \cite{he2016deep} & 60 & 3594 & 
			0.991±0.002 & 0.854±0.008 & 0.988±0.001 & 0.999±0.0 &
			0.792±0.032 & 0.663±0.017 & 0.724±0.01 & 0.876±0.052 &
			0.799±0.047 & 0.681±0.006 & 0.742±0.002 & 0.908±0.027  \\ \hline
			
			ResNet50 \cite{he2016deep} & 25 & 1197 & 
			0.98±0.002 & 0.856±0.003 & 0.974±0.006 & 0.998±0.0 &
			0.754±0.021 & 0.632±0.054 & 0.646±0.02 & 0.847±0.04 &
			0.828±0.002 & 0.698±0.005 & 0.747±0.003 & 0.896±0.017  \\ \hline
			
			ResNet101 \cite{he2016deep} & 44 & 2413 & 
			0.979±0.005 & 0.853±0.005 & 0.974±0.007 & 0.999±0.001 &
			0.743±0.079 & 0.617±0.05 & 0.65±0.07 & 0.845±0.038 &
			0.839±0.008 & 0.709±0.006 & 0.744±0.002 & 0.888±0.011  \\ \hline
			
			ResNet152V2 \cite{he2016identity} & 60 & 3481 & 
			0.982±0.006 & 0.856±0.01 & 0.978±0.006 & 0.999±0.0 &
			0.779±0.033 & 0.642±0.023 & 0.717±0.006 & 0.866±0.059 &
			0.856±0.01 & 0.704±0.01 & 0.782±0.01 & 0.871±0.01  \\ \hline
			
			ResNet101V2 \cite{he2016identity} & 44 & 2300 & 
			0.996±0.006 & 0.852±0.01 & 0.995±0.007 & 0.999±0.0 &
			0.819±0.008 & 0.688±0.003 & 0.731±0.002 & 0.851±0.027 &
			0.85±0.01 & 0.713±0.01 & 0.786±0.01 & 0.879±0.01 \\ \hline
			
			ResNet50V2 \cite{he2016identity} & 25 & 1085 & 
			0.99±0.006 & 0.863±0.01 & 0.987±0.006 & 0.999±0.0 &
			0.809±0.022 & 0.668±0.019 & 0.714±0.017 & 0.846±0.029 &
			0.86±0.01 & 0.703±0.01 & 0.754±0.01 & 0.874±0.01  \\ \hline
			
			InceptionV3 \cite{szegedy2016rethinking} & 23 & 586 & 
			0.979±0.005 & 0.871±0.01 & 0.975±0.004 & 0.998±0.001 &
			0.783±0.023 & 0.683±0.023 & 0.747±0.007 & 0.893±0.025 &
			0.87±0.01 & 0.745±0.01 & 0.815±0.01 & 0.898±0.01   \\ \hline
			
			MobileNet \cite{howard2017mobilenets} & 4 & 146 & 
			0.979±0.007 & 0.857±0.004 & 0.974±0.01 & 0.999±0.0 &
			0.734±0.172 & 0.508±0.161 & 0.463±0.158 & 0.6±0.101 &
			0.684±0.146 & 0.602±0.098 & 0.589±0.175 & 0.617±0.119 \\ \hline
			
			InceptionResNetV2 \cite{szegedy2017inception} & 55 & 1258 &
			0.997±0.009 & 0.849±0.01 & 0.994±0.007 & 0.999±0.001 &
			0.845±0.048 & 0.717±0.005 & 0.761±0.021 & 0.89±0.028 &
			0.864±0.01 & 0.759±0.01 & 0.813±0.01 & 0.907±0.01  \\ \hline
			
			Xception \cite{chollet2017xception} & 22 & 1424 &
			0.999±0.004 & 0.871±0.001 & 0.999±0.004 & 0.999±0.0 &
			0.889±0.026 & 0.763±0.021 & 0.839±0.025 & 0.874±0.036 &
			0.916±0.017 & 0.778±0.007 & 0.866±0.025 & 0.894±0.008 \\ \hline
			
			CU-Net \cite{feng2019cu} & 82 & 3372 & 
			0.999±0.009 & 0.903±0.01 & 0.999±0.009 & 0.999±0.0 &
			0.904±0.005 & 0.782±0.006 & 0.878±0.005 & 0.904±0.02 &
			0.895±0.01 & 0.807±0.01 & 0.856±0.01 & 0.897±0.01  \\ \hline
			
			\textbf{FabricNet(ours)} & \textbf{4.8} & \textbf{640} & 
			\textbf{0.999±0.009} & \textbf{0.879±0.01} & \textbf{0.999±0.009} & \textbf{0.999±0.0} &
			\textbf{0.929±0.001} & \textbf{0.855±0.001} & \textbf{0.922±0.002} & \textbf{0.948±0.023} &
			\textbf{0.912±0.01} & \textbf{0.846±0.01} & \textbf{0.902±0.01} & \textbf{0.932±0.01}  \\ \hline
		\end{tabular}
	\end{adjustbox}
\end{table}

The actual Xception architecture consists of one entry flow, nine middle flows, and an exit flow network. Each flow network contains residual connections. To select a proper head model, we further investigate the Xcecption architecture with different flow settings. Figure \ref{fig:xception_flows} represents a validation test score of the Xception architecture tuning the number of entry flow, middle flow, and exit flow segments. Only selecting the entry flow causes the validation F1-score and AUC to decrease. Choosing different numbers of middle flow layers improves the validation score by 0.2. However, for a different number of middle flow blocks, the score remains nearly constant. We select an entry flow with two middle flow network as the head architecture of the FabricNet. Although setting six intermediate flow blocks with an entry block acquires the highest score, the improvement is negotiable. As we implement a class-wise ensemble network that will contain more parameters (in the ensemble model), we avoid over-parameterizing the head model. Avoiding over-parameterization causes the overall architecture to be less prone to overfitting. Furthermore, using one entry and two middle blocks as the head model decreases the number of the trainable parameter by 80\% compared to Xception architecture. Also, reducing the number of parameters causes the reduction of FLOPs by 60\%, corresponding to Xception architecture.

\begin{figure}[H]
	\center
	\includegraphics[width=0.7\linewidth]{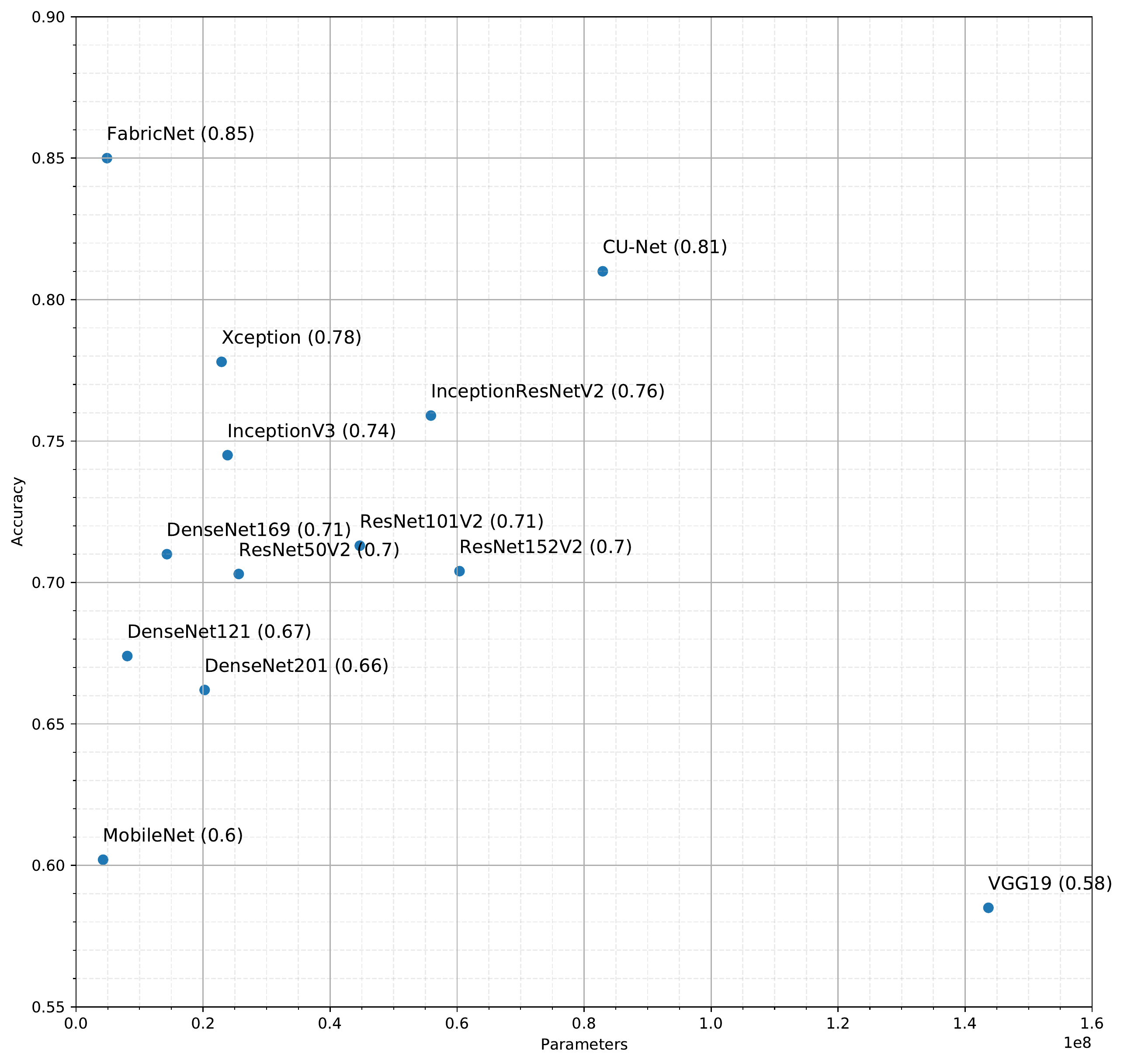}
	\caption{A scatter plot illustrating the test accuracy scores (vertical axis) w.r.t. the number of trainable parameters (horizontal axis). Zoom in for a better view.}
	\label{fig:scatter_plot}
\end{figure} 

We further investigate for selecting the optimal ensemble model. However, it is considered to keep the ensemble network's trainable parameters as less as possible to avoid overfitting and for easy training. Hence, we search for an optimal shallow architecture as an ensemble model. Shallow architecture requires fewer parameters, and as a result, it is possible to greatly increase the number of networks in the ensemble, based on the output classes. Therefore, we investigate CNN architectures with no more than three layers. Table \ref{tab:ensemble_archi} exhibits our experiment with different ensemble architecture with the reported F1-score on validation and test dataset. In the ensemble architecture, only separable convolution is implemented as it initiates fewer parameters. A close relationship between the trainable parameters and overfitting can be observed by investigating the table data. Higher training parameters in the ensemble model results in overfitting in the validation data. By decreasing the number of training parameters, a reduction in overfitting can be observed. However, after a certain period, the score drops. Adding additional layers does not heavily improve the score. Thus, each of the ensemble architecture is implemented using two depthwise seperable convolutions followed by a single fully-connected node. A sigmoid activation function is used as our target output may contain multiple classes at once.

Figure \ref{fig:fabricnet_flows} represents a comparison (on validation dataset) of FabricNet and the CU-net architecture. Both of the architectures follow ensemble strategy. However, the architectures' influential differences are: 1) improper class distribution for each ensemble model and 2) the number of parameters for each ensemble model. In the case of CU-Net architecture, the output of each class is generated based on the ensemble framework's decision. However, the architecture does not define specific ensemble models for each specific class. Hence, a single binary output of each class can be easily biased by multiple ensemble models. On the contrary, FabricNet architecture contains shallow CNN models specifically for each class. Hence, FabricNet avoids biased outputs for each class. Also, in comparison to CU-Net architecture, FabricNet requires a low number of trainable parameters. Lower parameters solve the overfitting issue, also models with lesser parameters are more comfortable to train.

The FabricNet architecture (network illustration in Figure \ref{fig:model}) is compared with the DCNN architectures, presented in Table \ref{tab:acc}. The comparison reports the precision, accuracy, F1-score, and AUC score of all the architectures calculated on the train, validation, and test dataset. The table represents the improvement of FabricNet architecture from the general implementation of Xception architecture. Although the FabricNet is a subset of the Xception architecture, joining the ensemble models boosts the F1-score of the FabricNet architecture by approximately 0.7. Figure \ref{fig:scatter_plot} represents a scatter plot of the tested architectures and the FabricNet. The horizontal axis indicates the number of training parameters, and the vertical axis reports the F1-score. As to calculate the number of trainable parameters and the FLOPs of the FabricNet architecture, the whole ensemble of 50 classifiers is considered. The FabricNet architecture achieves the highest accuracy while keeping the training parameters at a limit of 4.8 million. On the contrary, the MobileNet and MobileNetV2 architectures fall behind in achieving a better score with comparable training parameters. It is a clear indication that FabricNet architecture gains superiority due to the CNN network topology. 

Residual architectures (ResNet, Xception, Inception) indirectly implements the property of the ensemble strategy. Residuals not only help to solve the vanishing gradient problem but also can ignore a particular CNN block if required. However, the ignore state of a CNN block depends on the type of input, and it is adequately utilized using backpropagation. Yet, the difference between the residual ensemble and our implemented ensemble lies in the dedicated path (i.e., a layer sequence is fully dedicated to a class). Therefore each of the filters only looks for class-specific features. The dedicated feature extraction solves separating the depth filters for a class-specific identity extraction mostly occurred at the deepest layers (before fully connected layer) of a CNN architecture.

\begin{figure}[H]
	\center
	\includegraphics[width=\linewidth]{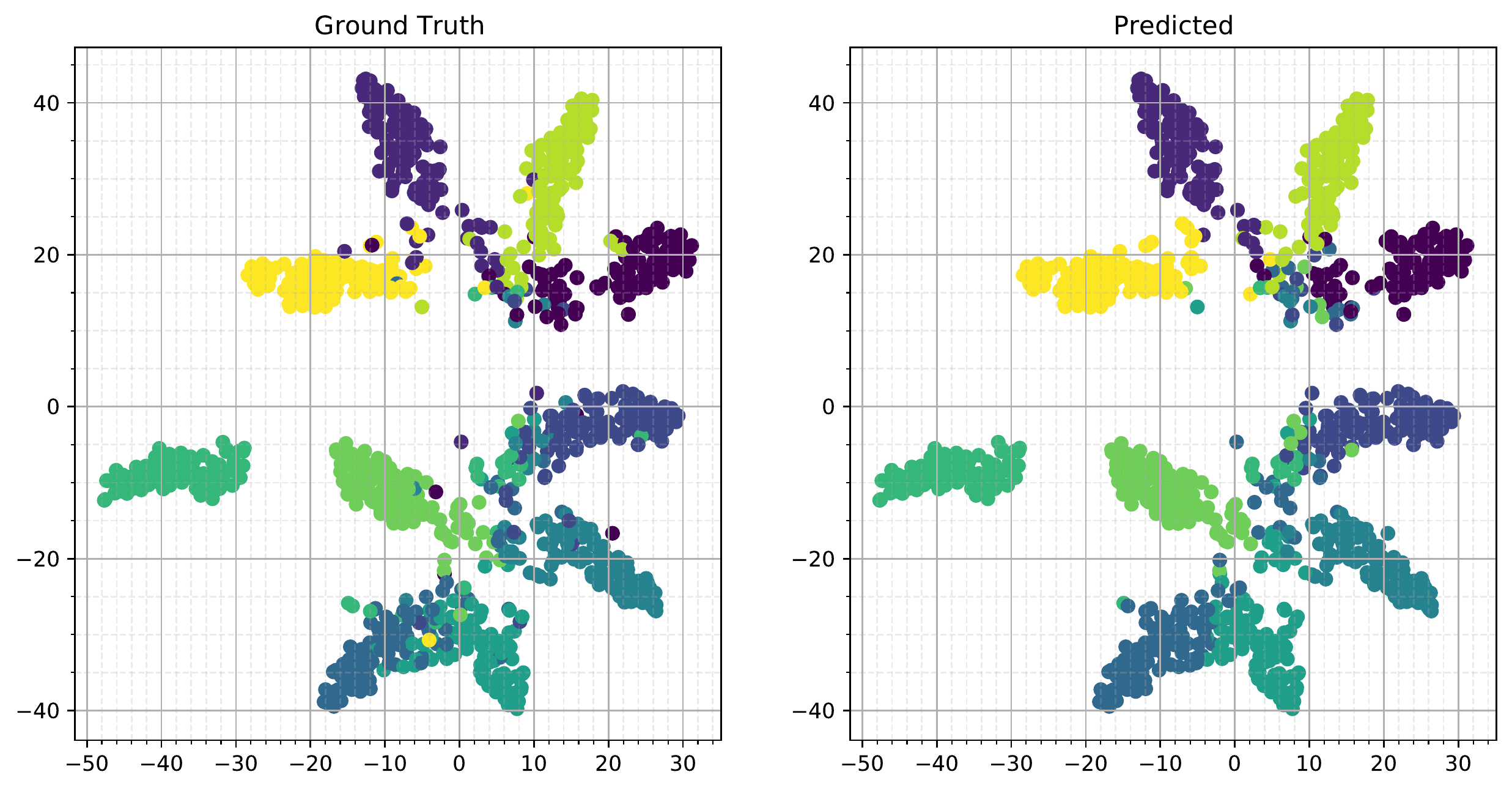}
	\caption{The figures illustrate scatter-plots generated by the head model of FabricNet, on a small portion of the dataset (with ten classes). The head model tries to assume the classes, whereas the ensembles try to memorize the head ensemble's missed out features. Mixing the head and ensemble model enables the architecture to perform better with more class-specific feature memorization. }
	\label{fig:scatter_plot2}
\end{figure} 

Figure \ref{fig:scatter_plot2} illustrates the embedding vector outputs generated by the head model of the FabricNet architecture on a small subset of the input. It can be anticipated by analyzing the figure that the head model extricates some of the necessary features. Further, the ensemble's class-specific models furnish the embeddings based on the class-specific features. In such a case, the ensemble has the primary advantage of correcting the head model's erroneous guesses and further fix the issue through memorization.

\section{Conclusion}
\label{sec:conclusion}

The paper presents an architecture FabricNet, a textile fiber recognition scheme, that can recognize multiple fibers at once by only processing the surface image of fabrics. This research work points to an immense improvement in fiber recognition tasks as the previous methods required microscopic images and spectrographs of fibers. The FabricNet is implemented based on a new idea of ensemble architecture, and to outline the difference, the paper comprises an investigation of mostly implemented ensemble architectures. The experiment is conducted using fifty types of textile fibers, and the FabricNet outperforms most of the well-known image classification architectures. We strongly believe that the overall contribution of this paper inaugurates a broader perception in the scope of image pattern recognition and industrial fiber identification research works.

\bibliographystyle{unsrt}
\bibliography{references.bib}
\end{document}